\documentclass{article} 
\usepackage[final]{colm2026_conference}

\usepackage{microtype}
\usepackage{hyperref}
\usepackage{url}
\usepackage{booktabs}
\usepackage{graphicx}
\usepackage{subcaption}
\usepackage{amsmath} 
\usepackage{wrapfig}
\usepackage{soul}
\usepackage{fontawesome5}

\usepackage{lineno}

\definecolor{darkblue}{rgb}{0, 0, 0.5}
\hypersetup{colorlinks=true, citecolor=darkblue, linkcolor=darkblue, urlcolor=darkblue}

\title{Loop, Think, \& Generalize: Implicit Reasoning in Recurrent-Depth Transformers}

\author{
Harsh Kohli \hspace{1.1em} Srinivasan Parthasarathy \hspace{1.1em} Huan Sun\thanks{Equal supervision.} \hspace{1.1em} Yuekun Yao\footnotemark[1] \\
The Ohio State University\\
\texttt{\{kohli.120,parthasarathy.2,sun.397,yao.1267\}@osu.edu} \\ [0.2em]
\small \faGithub\ \url{https://github.com/OSU-NLP-Group/Loop-Think-Generalize}
}

\begin{document}

\ifcolmsubmission
\linenumbers
\fi

\maketitle

\begin{abstract}
We study implicit reasoning, i.e.\ the ability to combine knowledge or rules within a single forward pass.
While transformer-based large language models store substantial factual knowledge and rules, they often fail to compose this knowledge for implicit multi-hop reasoning, suggesting a lack of compositional generalization over their parametric knowledge.
To address this limitation, we study recurrent-depth transformers, which enables iterative computation over the same transformer layers.
We investigate two compositional generalization challenges under the implicit reasoning scenario: \textit{systematic generalization}, i.e.\ combining knowledge that is never used for compositions during training, and \textit{depth extrapolation}, i.e.\ generalizing from limited reasoning depth (e.g.\ training on up to 5-hop) to deeper compositions (e.g.\ 10-hop).
Through controlled studies with models trained from scratch, we show that while vanilla transformers struggle with both generalization challenges, recurrent-depth transformers can effectively make such generalization.
For systematic generalization, we find that this ability emerges through a three-stage grokking process, transitioning from memorization to in-distribution generalization and finally to systematic generalization, supported by mechanistic analysis.
For depth extrapolation, we show that generalization beyond training depth can be unlocked by scaling inference-time recurrence, with more iterations enabling deeper reasoning.
We further study how training strategies affect extrapolation, providing guidance on training recurrent-depth transformers, and identify a key limitation, \textit{overthinking}, where excessive recurrence degrades predictions and limits generalization to very deep compositions.
\end{abstract}

\section{Introduction}
Large language models (LLMs) \citep{brown2020language} are known to acquire substantial factual knowledge during pretraining, storing it in their parameters \citep{geva-etal-2023-dissecting}. 
However, how effectively this knowledge can be composed for reasoning remains less understood \citep{10.5555/3666122.3669203, press-etal-2023-measuring}.
In particular, recent work shows that transformer-based LLMs struggle under \textbf{implicit reasoning}, i.e.\ reasoning within a \textit{single forward pass} without explicit chain-of-thought (CoT) \citep{wei2022chain}. 
Such failures reveal a fundamental limitation of transformers: despite storing rich knowledge, they are often unable to flexibly combine it to solve novel questions.
This limitation has important implications for generalization, as many tasks require composing multiple pieces of seen knowledge in novel ways not observed during training \citep{lake2018generalization, berglund2023reversal}.

Why do transformers struggle to combine their parametric knowledge in implicit reasoning? 
Consider a query such as \textit{“The spouse of the performer of Imagine is”}. 
Previous work shows that transformers solve this by chaining two facts: first retrieving that \textit{the performer of Imagine is John Lennon} in shallow layers, and then that \textit{the spouse of John Lennon is Yoko Ono} in deeper layers \citep{biran-etal-2024-hopping, 10.5555/3737916.3740933, yang-etal-2024-large-language-models}. 
However, since knowledge is distributed across different layers of the transformer, there is no guarantee that the fact required for a particular query can be accessed correctly.
For example, if the fact \textit{the spouse of John Lennon is Yoko Ono} is only stored in shallow layers, deeper layers cannot access it because parameters are not shared across layers. 
While transformers can be trained to learn to combine such knowledge properly \citep{10.5555/3737916.3740933, yao-etal-2025-language-models}, they fail to compositionally generalize to unfamiliar combinations or deeper recursive combinations.

To address this limitation, we introduce depth-recurrence into transformers, allowing the same set of layers to be applied iteratively. The input sequence is processed multiple times by a shared transformer block, where the output of each iteration serves as input to the next. 
In contrast to vanilla transformers, where knowledge is tied to specific layers, recurrence enables more flexible access to and composition of parametric knowledge within a single forward process.
Such models, known as \textit{recurrent-depth transformers} or \textit{looped transformers}, have recently gained attention as a promising architecture \citep{dehghani2018universal, geiping2025scaling, zhu2025scaling}.
While prior work has shown that recurrent-depth transformers improve length generalization \citep{bansal2022end, fan2024looped}, it remains unclear whether they can overcome compositional generalization limitations when reasoning over parametric knowledge.

In this paper, we systematically study whether recurrent-depth transformers can compositionally combine their parametric knowledge \textit{implicitly}. 
By constructing synthetic datasets, we train models to learn implicit reasoning from scratch. Unlike LLMs trained on vast, opaque web-scale corpora, this setup provides control over the data and mitigates confounding biases introduced during pretraining.
Specifically, we characterize two challenges: \textit{systematic generalization} (combining knowledge not used in any composition during training) and \textit{depth extrapolation} (e.g., training on 5-hop reasoning and evaluating on 10-hop).

Our main findings are two-fold. 
First, \textbf{recurrent-depth transformers exhibit strong systematic generalization, while vanilla transformers fail to do so}. 
We show that this ability emerges through a sharp three-stage \textit{grokking} process, 
that transitions from memorization to in-distribution generalization, and finally to systematic generalization.
We also support this with evidence from the internal activations of models across different training stages.

Second, \textbf{recurrent-depth transformers enable depth extrapolation}, generalizing to reasoning depths beyond those observed during training, as inference-time compute (i.e., recurrent iterations) increases. 
We further find that the training-time recurrence strategy plays a critical role in extrapolation performance, with dynamic recurrence achieving the strongest generalization. 
Despite these gains, we identify a key limitation: recurrent-depth transformers suffer from overthinking \citep{bansal2022end}, which degrades performance and limits generalization to extremely deep recursions. 

\begin{figure*}[t]
  \centering
  \includegraphics[width=\linewidth]{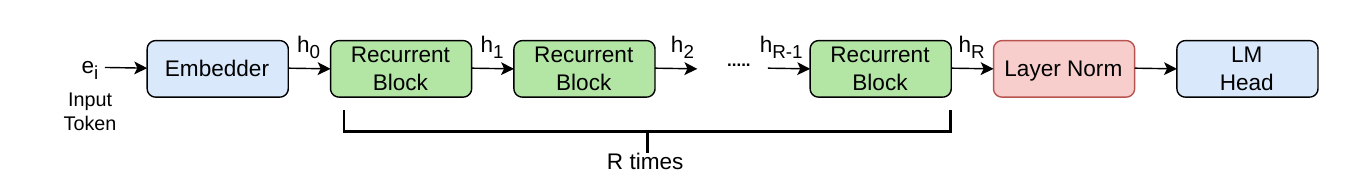}
  \caption{\textbf{Recurrent depth model architecture.} The transformer block is repeated $R$ times. The embedding layer and language model head (LM Head) have tied weights. In our experiments, we use a simple looped transformer similar to \citet{saunshireasoning} without design elements such as input injection, gated halting, and middle looping.}
  \label{fig:model_diagram}
\end{figure*}

\section{Related Work}

Several small-scale studies pretrain looped or recurrent-depth transformers on synthetic tasks to better understand their behavior in a controlled setting. Our work best aligns with such studies where we are able to cleanly attribute differences in performance and generalization to specific architectural choices and model design decisions. \citet{yang2024looped} demonstrate how "looping" a transformer block helps to better emulate learning algorithms such as gradient descent for in-context linear regressions, 2-layer neural networks, and decision trees. \citet{fan2024looped} show that such looped transformers offer superior length generalization on algorithmic tasks such as parity and binary addition. \citet{saunshireasoning} conduct a larger-scale pretraining with the 250B tokens of the Pile dataset \citep{gao2020pile} and find that looped versions of transformer models of the same effective depth have a greater inductive bias towards reasoning at the cost of memorization and perplexity.  Based on these results, they propose a regularization term that encourages certain layers to be closer to each other, thus improving the tradeoff between reasoning and fact recall.  

Relative to other works, our targeted setting yields unique insights on training dynamics and model behavior.
We demonstrate how weight sharing through recurrence can solve systematic composition where vanilla transformers are known to struggle and extrapolation in multi-hop composition is possible with increased recurrence at inference-time. While \citet{fan2024looped} propose looped architectures for length generalization, they assume an oracle number of training iterations based on sample complexity. 
We believe that our setup is closer to real-world scenarios where task complexity cannot be easily estimated through heuristics (such as input length). 
Without the assumption of task complexity a priori, we face distinct challenges in training our models. 
We analyze how best to apply methods like recurrent-depth and common pitfalls to avoid, which can help inform more robust implicit reasoning models in the future. We discuss other related work in Appendix~\ref{app:other_related_work}.

\section{Task Formulation}



\begin{wrapfigure}{r}{0.55\linewidth}
    \centering
    \vspace{-2.0em}
    \includegraphics[width=\linewidth]{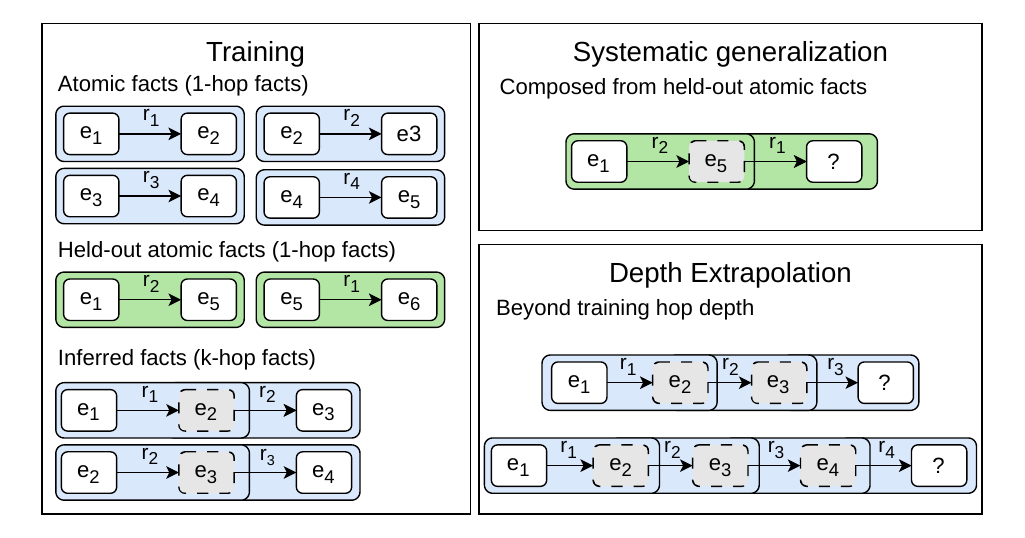}
    \caption{Illustration of systematic and extrapolation generalization tasks with a sample dataset.}
    \label{fig:dataset-creation}
    \vspace{-1.0em}
\end{wrapfigure}

We formally define our implicit reasoning setup using a synthetic multi-hop reasoning task, and categorize three generalization challenges under this formulation: in-distribution generalization, systematic generalization, and depth extrapolation (Figure \ref{fig:dataset-creation}).
The latter two can be viewed as out-of-distribution (OOD) generalization. Such tasks have been shown to be difficult for vanilla transformers to learn \citep{yao-etal-2025-language-models}, highlighting their limitations in composing parametric knowledge for reasoning \citep{allen2023physics, yang-etal-2024-large-language-models}.


\subsection{Task Definition}
Our implicit reasoning task relies on a directed knowledge graph (KG) where nodes represent a set of entities $E = \{e_i\}$ and edges represent a set of relations $R = \{r_j\}$. 
The KG is composed of atomic (1-hop) facts, each taking the form of a triplet $(h, r, t)$, where $h, t \in E$ and $r \in R$ (here $h$ and $t$ imply the head and tail entities, respectively).

A $k$-hop inferred fact is defined as a chain of $k$ atomic facts connecting a head entity $h$ to a final tail entity $t$ via a sequence of $k-1$ intermediate entities ($i_1, \dots, i_{k-1}$):
$$(h, r_1, i_1), (i_1, r_2, i_2), \dots, (i_{k-1}, r_k, t)$$
    Given the head entity $h$ and the sequence of relations $r_1, \dots, r_k$, we use an auto-regressive decoder-only model to predict the final tail entity $t$. 
    The input prefix is $<e_h><r_1><r_2>\dots<r_k>$, and the target is $<e_t>$.
    Ideally, the model must implicitly perform the $k$-hop traversal, successively retrieving each intermediate entity ($i_1, \dots, i_{k-1}$) until it can resolve the final tail entity $t$.

\subsection{Generalization Challenges}

Given a generated knowledge graph, we first define the complete atomic fact set as $\mathcal{C} = \{(h, r, t)\}$, and the induced set of $k$-hop inferred facts from $\mathcal{C}$ as
\[
\mathcal{I}_k(\mathcal{C}) = \{(h, r_1, \dots, r_k, t)\ \mid\ \exists\ i_1, \dots, i_{k-1},\ (h, r_1, i_1), \dots, (i_{k-1}, r_k, t) \in \mathcal{C} \}.
\]
\paragraph{Training set.} The training set includes two parts: all possible atomic facts $\mathcal{C}$ together with a set of inferred facts $\mathcal{I}_{train}$ up to a maximum depth $k_{train}$ (e.g. $k$-hop facts with $k \in [2, k_{train}]$).
To characterize different generalization challenges, we partition the atomic fact set into two disjoint subsets
$\mathcal{C} = \mathcal{C}_{ID} \cup \mathcal{C}_{OOD}$.
The training inferred facts $\mathcal{I}_{train}$ can then be defined as
\[
\mathcal{I}_{train} = \{(h, r_1, \dots, r_k, t)\ \mid\ (h, r_1, \dots, r_k, t) \in \mathcal{I}_k(\mathcal{C}_{ID}),\ k \le k_{\text{train}} \}.
\]

We then define three generalization challenges:
\paragraph{In-distribution generalization.}
The model is evaluated on inferred facts $(h, r_1, \dots, r_k, t) \in \mathcal{I}_k(\mathcal{C}_{ID})$ that are not observed during training, equivalent to randomly sample inferred facts from $\mathcal{I}_k(\mathcal{C}_{ID})$ as held-out test set. 
Despite its simplicity, previous work shows that vanilla transformers can only learn such tasks through extended training \citep{10.5555/3737916.3740933}.

\paragraph{Systematic generalization.}
The model is evaluated on inferred facts $(h, r_1, \dots, r_k, t) \in \mathcal{I}_k(\mathcal{C}_{OOD})$, which are induced from atomic facts that are never used in compositions in the training data. 
This requires the learner to systematically combine its learned knowledge, without having seen combinations of it in training. 
This setting simulates scenarios where knowledge appears only as plain text in pretraining data (e.g.\ long-tail knowledge), but never forms answers to reasoning queries during training. 
Previous work \citep{10.5555/3737916.3740933} shows that vanilla transformers completely fail on this generalization challenge.

\paragraph{Depth extrapolation.}
The model is evaluated on inferred facts of greater depth than those included in the training dataset, i.e., $(h, r_1, \dots, r_k, t) \in \mathcal{I}_k(\mathcal{C_{ID}})$ with $k$ larger than $k_{train}$.
Solving this requires the learner to infer the underlying rules of the task and iteratively apply them at depths far beyond those observed during training. 
This setting simulates scenarios where the complexity (i.e.\ reasoning depth) of training data is limited due to budget constraints, yet we expect the model to generalize beyond training.
Such depth generalization poses challenges for vanilla transformers in symbolic \citep{kim-linzen-2020-cogs} and knowledge reasoning \citep{yao-etal-2025-language-models}.
Although related to length generalization, depth extrapolation is conceptually distinct: it measures the depth to which a model can repeatedly apply learned rules over its parametric knowledge beyond training.

\section{Recurrent-Depth Transformer}
\label{sec:model-design}
\paragraph{Model architecture.}
Across all experiments we use a decoder-only transformer with a \emph{recurrent-depth} design illustrated in Figure~\ref{fig:model_diagram}. Concretely, we instantiate a GPT-2 style block with $L$ layers and reuse this for $R$ recurrent iterations, yielding an effective rolled-out depth of $D = L \times R$ layers. At each recurrent iteration the same stack of layers is applied to the current hidden states. This allows the model to allocate more computation (increasing $R$) at inference time without changing the architecture or re-training the parameters. We exploit this property in our inference-time scaling experiments described in Section~\ref{sec:multihop}. Formally, let $f_\theta$ denote the $L$ transformer layers with shared parameters~$\theta$, and let $h^{(0)}$ be the input sequence after the initial embedding layer. The model computes
\[
    h^{(r+1)} = f_\theta\big(h^{(r)}; m\big), \qquad r = 0, \dots, R-1,
\]
where $m$ denotes the causal attention and padding masks. The final representation $h^{(R)}$ is passed through a final layer normalization and a tied output projection to produce logits over the vocabulary at each position. We only supervise the next-token distribution at the final position corresponding to the tail entity $t$. Each entity and relation is represented by a dedicated token (\texttt{$<e_i>$}, \texttt{$<r_j>$}), and the query prefix is mapped to token embeddings. 

We adopt a zero-initialization strategy to stabilize training under repeated application of shared weights.
Specifically, we initialize the output projection matrices (\texttt{c\_proj}) of both the multi-head attention and feed-forward blocks to zero, so that each recurrent block is an exact identity mapping at initialization.
This ensures that the input-output Jacobian remains stable even when the model is unrolled to a large number of recurrent iterations.
This design is motivated by the known instability of deep networks with shared parameters \citep{agarwala2022deep}, which becomes particularly pronounced in recurrent-depth transformers \citep{saunshireasoning}.
Following \citet{zhang2019fixupinitializationresiduallearning}, this initialization supports stable optimization under unbounded unrolling of the recurrent iterations.


\paragraph{Stopping strategies of the recurrent iterations.} Training the looped transformer requires a stopping strategy to determine the number of recurrent iterations in the forward pass on the input. 
We consider two stopping strategies: \textit{fixed iteration} and \textit{dynamic iteration}. 
Fixed iteration determines the recurrent iterations to be the same fixed value for all training instances.
The dynamic iteration strategy samples the number of recurrent iterations independently for each training batch. Concretely, for the dynamic model we sample
\[
R \sim \mathrm{clip}\big(\mathrm{Poisson}(\lambda), R_{\min}, R_{\max}\big),
\]
Unless otherwise stated, we set the hyperparameters to \(R_{\min}=2\) and \(R_{\max}=8\).
Such strategies have been shown to be effective in realistic pretraining \citep{geiping2025scaling, zhu2025scaling}, and here we adopt a simple Poisson distribution to control the sampling distribution.
Importantly, our strategies contrast with prior studies on the generalization ability of looped transformers, where the iteration number is matched to the complexity of each training instance, assuming oracle access to such complexity.
Instead, our setup reflects practical scenarios \citep{geiping2025scaling}, where the complexity is unknown and computation cannot be allocated precisely in advance.

\section{Systematic Generalization}\label{sec:systematicity}
In this section, we study systematic generalization, i.e.\ whether models can combine parametric knowledge not composed during training for multi-hop tasks. 
We focus on $2$-hop, as \citet{10.5555/3737916.3740933} shows that vanilla transformers already struggle with this simple task.
\subsection{Experiment Setup} \label{sec:systematicity:setup}
\paragraph{Dataset.} We construct the dataset by instantiating a knowledge graph with $|E|=2000$ and $|R|=200$, where each entity has average out-degree 20. 
We then include all 40k atomic facts, randomly partitioned with 95\% $C_{ID}$, and 5\% $C_{OOD}$, together with 273.6k inferred facts for training. 
Our in-distribution set includes 3k held-out two-hop inferred facts composed from $C_{ID}$, and OOD set includes nearly 2k two-hop inferred facts composed from $C_{OOD}$.

\paragraph{Model.} We train our looped transformer with $L=4$ layers, and use fixed training recurrence with $R \in \{1, 2, 4, 8\}$.
The model with $R=1$ is equivalent to a 4-layer vanilla transformer. 
We evaluate the accuracy of the predicted token against the gold answer.
Absolute position embeddings (APE) \citep{NIPS2017_3f5ee243} are used as positional embeddings in this setup. We do not use dynamic recurrence in this setting, as systematic generalization in the 2-hop task already emerges from weight sharing under fixed recurrence. In the multi-hop setting, however, it improves extrapolation to more complex samples at inference time and helps alleviate latent overthinking, as discussed in Section~\ref{sec:multihop}.

\subsection{Results}
\label{sec:sys_results}
\begin{figure}[h]
    \centering
    \includegraphics[width=\columnwidth]{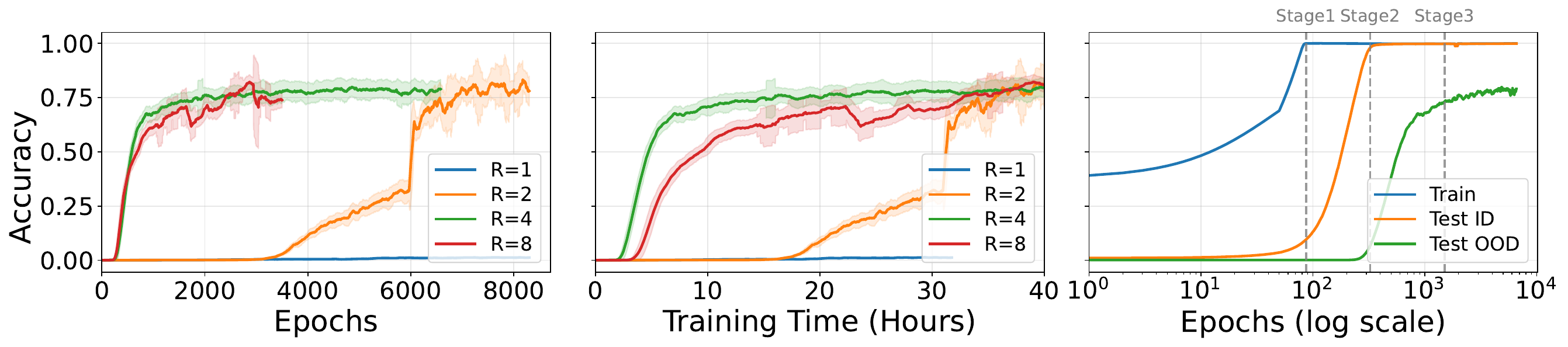}
    
    \caption{
    Accuracy curves for recurrent-depth models across training epochs and wall-clock time. 
    \textbf{Left:} Test OOD accuracy for models trained with $R \in \{1,2,4,8\}$, plotted against training epochs. Curves are smoothed with a 100-epoch rolling mean, with shading indicating standard deviation.
    \textbf{Middle:} Test OOD accuracy for the same models, plotted against training wall-clock time (hours). 
    \textbf{Right:} Accuracy of the $R=4$ model on the examples from training, ID, and OOD test splits, plotted against training epochs. 
}
    \label{fig:composition-test-ood}
\end{figure}

\paragraph{Recurrent-depth transformers perform systematic generalization, while vanilla transformers do not.}
On the left of Figure~\ref{fig:composition-test-ood}, we plot OOD accuracy as a function of training epochs.
We find that the vanilla transformer (i.e.\ $R=1$) completely fails when the task requires combining unfamiliar atomic facts, while even the simplest $R=2$ recurrence achieves non-trivial generalization performance.
Increasing training iterations further accelerates the convergence, e.g.\, $R=4$ converges with 2k epoch, while $R=2$ takes 7k. This acceleration is not only in terms of training steps, but also in absolute wall-clock time (Figure~\ref{fig:composition-test-ood}, middle).

\paragraph{Systematic generalization emerges through a three-stage grokking dynamic.}
We further analyze the training dynamics of the $R=4$ model (right panel of Figure~\ref{fig:composition-test-ood}) to understand how systematic generalization emerges.
We observe a three-stage dynamic:
In the first stage, the model overfits the training set, with only training accuracy improving.
In the second stage, in-distribution generalization emerges after prolonged training beyond memorization, a phenomenon referred to as \textit{grokking}.
In the final stage, systematic generalization arises only after the model achieves near-perfect in-distribution accuracy, occurring at a much later point than training overfitting (e.g., $10^4$ vs.\ $10^2$ epochs).

\begin{figure}[h] 
\centering \includegraphics[width=\linewidth]{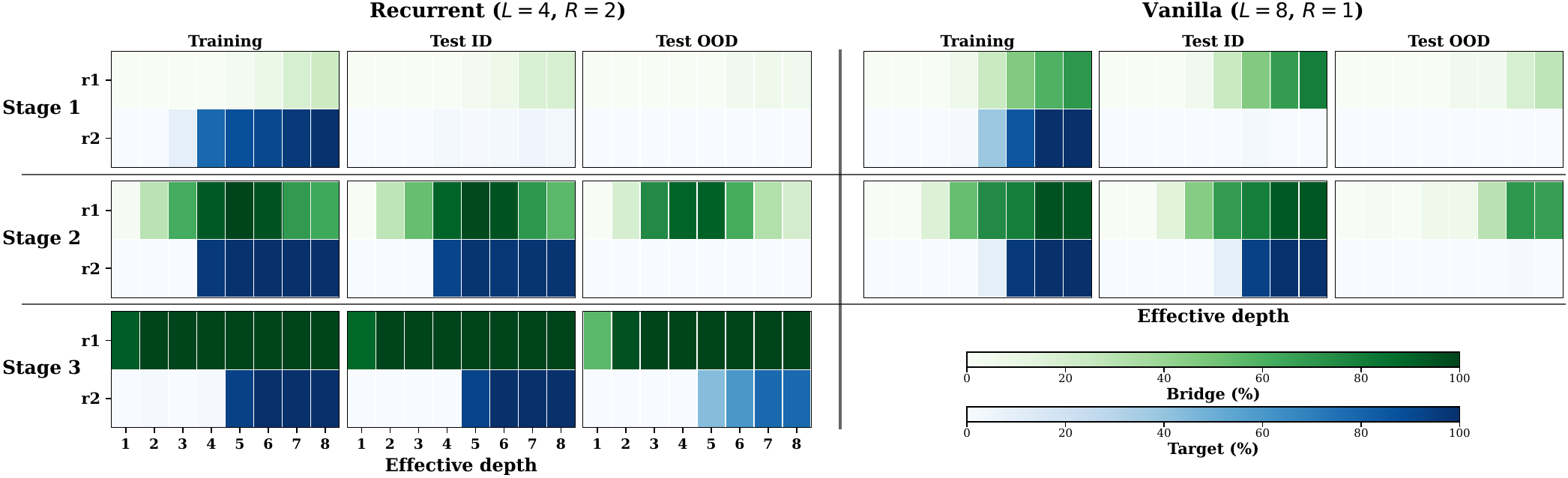} 
\caption{Accuracy of predicting bridge and target entities using logit lens at corresponding token positions for the recurrent-depth ($R=2$) and the 8-layer vanilla transformer.} 
\label{fig:heatmap1} 
\end{figure}

\paragraph{Analyzing model internals with logit lens.}
We use the logit lens technique \citep{nostalgebraist2020logitlens} to examine how models represent the bridge entity and the final target during different stages of training. After each layer and recurrent iteration, we project the intermediate hidden states through the final layer norm and language modeling head to obtain logits over the output vocabulary. For 2-hop inputs of the form $(h, r_1, r_2)$, where $h$ is the head entity and $r_1, r_2$ are the two relations, we measure at each effective depth the accuracy of predicting the bridge entity at the $r_1$ position and the target entity at the $r_2$ position. 
Figure~\ref{fig:heatmap1} shows logits lens for an $R=2$ recurrent-depth model on Training, Test ID, and Test OOD splits, across checkpoints corresponding to the three training stages.
We compare against an iso-FLOP 8-layer vanilla transformer with matched effective depth.
The vanilla model exhibits only two training stages and fails to achieve non-zero systematic generalization regardless of training time, consistent with \citet{10.5555/3737916.3740933}. We also include a causal analysis for systematic generalization in Appendix~\ref{app:causal}.


\paragraph{Grokking marks a transition from memorization to systematic generalization.} 
We first focus on the recurrent-depth model (Figure~\ref{fig:heatmap1}  left panels), which exhibits distinct mechanisms across the three stages.
In Stage 1, the model predicts targets without reliably decoding the bridge, indicating memorization.
In Stage 2, the bridge becomes decodable, followed by correct target prediction for in-distribution data at deeper effective depths.
Only in Stage 3 does the model succeed on OOD inputs, marking a transition from rote learning to systematic composition.
In contrast, although vanilla transformers (right panels) can recover the bridge entity on Test OOD inputs, they fail to perform the second-hop reasoning, as they lack incentives to encode OOD facts in deeper layers.

\section{Depth Extrapolation}
In this section, we study depth extrapolation, i.e., whether the model can perform deeper recursions when combining its parametric knowledge than those observed during training. 

\label{sec:multihop}

\begin{figure}[b]
    \centering
    \includegraphics[width=\linewidth]{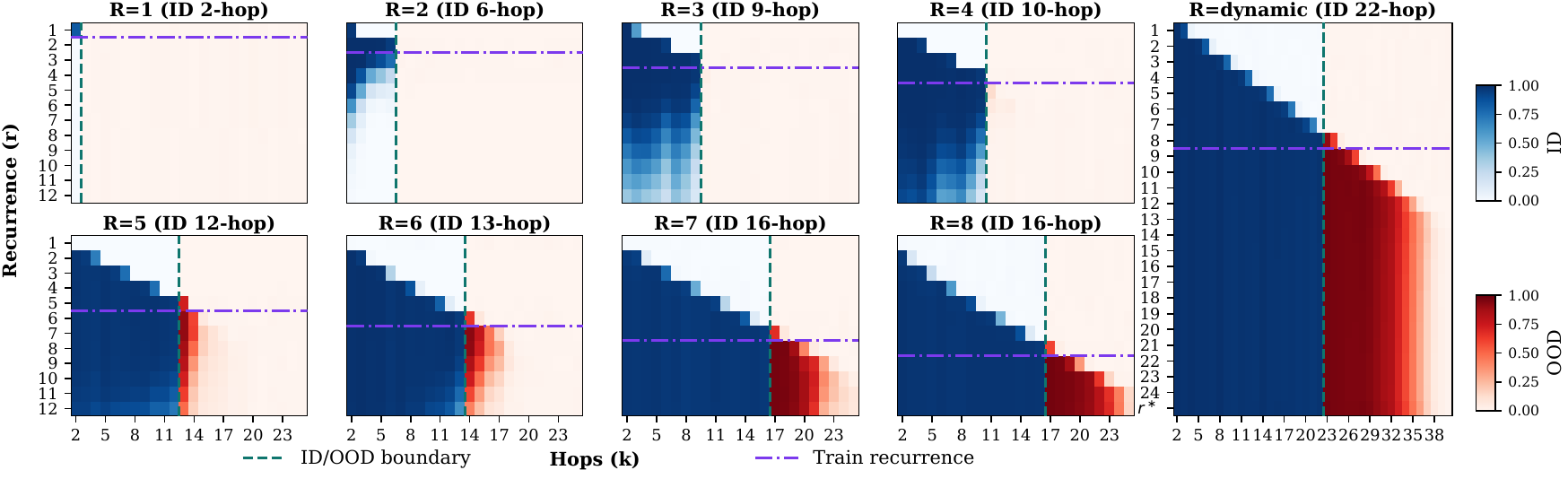}
\caption{Accuracy of recurrent-depth models on multi-hop composition, trained under various fixed and dynamic recurrence setups. The violet dash-dotted horizontal line indicates the training recurrence (or maximum in the dynamic setting), while the teal dashed vertical line marks the maximum ID generalization achieved by each model. The x-axis shows hop complexity and the y-axis shows inference-time recurrence ($r^*$ denotes adaptive halting).}
    \label{fig:extrapolation}
\end{figure}

\subsection{Experiment Setup}
\label{sec:exp_setup}

\paragraph{Curriculum training.}
Different from 2-hop scenarios, learning $k$-hop tasks generally requires training the model with an easy-to-hard curriculum over hop depth ($k$) as suggested in \citet{yao-etal-2025-language-models}. 
Specifically, we start with training on atomic and 2-hop facts until an accuracy threshold ($95\%$) is reached on a held-out 2-hop test split. 
Next, 3-hop data is included in the training for the next stage of our curriculum until $95\%$ is achieved on the held-out 3-hop test split. 
This process is repeated for each hop level $k \in \{2, \dots, N_{\max}\}$. 
To prevent forgetting, at each stage we jointly train on all previously introduced facts (e.g., at stage $k=5$ we train on atomic and 2-hop through 5-hop facts).

Due to this threshold-based curriculum, training facts beyond the model’s capability are never exposed.
That is, if the model fails to achieve above-threshold accuracy on $k$-hop queries, training terminates and $(k+1)$-hop data is never introduced.
We define the largest such $k$ as the \textit{learnable recursion depth} of the model.

\paragraph{Dataset.} 
We construct the dataset by instantiating a knowledge graph with $|E|=200$ entities and $|R|=10$ relations, where each entity has an average out-degree of 10. 
We additionally impose a permutation constraint on the knowledge graph to avoid learning shortcut solutions, for which we provide details in Appendix~\ref{app:permutation}. 
We pre-generate 2k atomic facts and 15k $k$-hop inferred facts for each $k \in [2, K_{max}]$, with $K_{max}=40$. 
During training, these facts are progressively introduced following the curriculum described above.

For each model, we evaluate on 750 held-out $k$-hop facts.
Facts with $k$ up to the model’s learnable recursion depth form the in-distribution test set, while those beyond it constitute the extrapolation test set.
This split is model-dependent, reflecting each model’s maximum achievable reasoning depth.


\paragraph{Model.} 
We follow the model setups in Section \ref{sec:systematicity:setup}, except that we use $R \in $\{1,2,3,4,5,6,7,8\} for fixed iteration.
Here we use no positional embeddings (NoPE) \citep{kazemnejad2023impact, wang2024length}, which shows better generalization in pilot studies. 
In addition to results with $L=4$, we present results with varying model size and dynamic train recurrence in Appendix~\ref{app:ratios}, and across random seed initializations in Appendix~\ref{app:seeds}. We also experiment with larger vanilla transformer models, iso-FLOP with recurrent-depth models using \(R \in \{2,4,6\}\) and report results in Appendix~\ref{app:isoflop_vanilla}.

\subsection{In-Distribution Generalization} \label{sec: extrapolation_id}
\paragraph{Scaling up training-time iteration increases the learnable recursion depth of looped transformers.} 
Looking at the blue area of Figure \ref{fig:extrapolation}, we find that increasing training recurrent iterations accordingly improves the ID generalization. 
This is consistent with previous findings \citep{10.5555/3737916.3740933,yao-etal-2025-language-models} that the learnable recursion depth of a transformer is bounded by the depth of its layers, 
and we demonstrate that for looped transformers, scaling up training recurrent iterations can also increase its "effective depth", without relying on additional parameters.
Training with dynamic iteration further increases the learnable recursion depth over the fixed iteration, suggesting that the fixed iteration is not an optimal design choice.
Interestingly, more recurrent iterations do not always translate into larger learnable depth. (e.g., both R=7 and R=8 learns up to 16-hop task).

\begin{wrapfigure}{r}{0.35\textwidth}
    \vspace{-0.8em}
    \centering
    \includegraphics[width=0.33\textwidth]{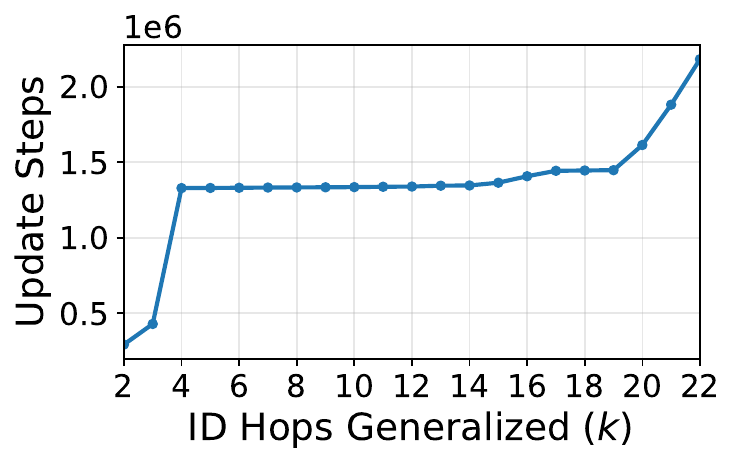}
    \vspace{-0.6em}
    \caption{Cumulative gradient updates required to first generalize to each hop complexity.}
    \label{fig:curriculum-cumulative}
    \vspace{-1.0em}
\end{wrapfigure}

\paragraph{Phase transition from prolonged training to rapid learning.}
We observe that models require a prolonged training phase to acquire low-hop tasks, after which they rapidly generalize to much more complex samples (Figure~\ref{fig:curriculum-cumulative}). 
We illustrate this for the model trained with dynamic recurrence by plotting training steps against the compositional complexity achieved on in-distribution data.
This suggests that the main difficulty lies in discovering the underlying compositional rule. 
Once such a rule is internalized, the model can quickly extend it to samples of much higher complexity. 
In Figure~\ref{fig:curriculum-cumulative} we observe how the model required over 1.3 million steps for grokking up to 4-hop train samples but was quickly able to learn up to 19-hop samples very few additional steps. Beyond that, for even more complex samples, while each new hop requires additional training steps to cross the 95\% threshold, the model still achieves strong generalization (\textgreater90\%) on hops 20, 21, and 22 within fewer than 8k extra steps per hop which is commensurate with the steps required for each additional hop from 4 through 19. By loading a trained checkpoint (20, say) and continuing training exclusively on 21-hop samples instead of the data mix consisting of samples from all previous stages in our curriculum, generalization over 90\% on the new split can be achieved in as little as 50 additional steps of training.

\subsection{Depth Extrapolation}
\paragraph{Scaling inference-time iterations unlocks depth extrapolation.}
In Figure~\ref{fig:extrapolation}, we observe that when using the same number of recurrent iterations as in training, all models struggle to generalize to tasks of higher complexity than those seen during training. 
However, this limitation is immediately alleviated when we increase the number of inference-time iterations, with more iterations enabling generalization to progressively harder tasks. 
Notably, this scaling effect only emerges for $R > 4$, suggesting that sufficient training-time iterations are a prerequisite for benefiting from increased inference-time computation.

\paragraph{The effect of training iteration strategy on extrapolation.}
The above results characterize the maximum reasoning depth each model can achieve under the curriculum setting,
but they do not disentangle whether the differences (e.g.\ $R=6$ generalizing to $17$-hop vs.\ $R=8$ to $24$-hop) are due to more training iterations or exposure to more complex training data (e.g.\ $R=6$ is trained up to $13$-hop, while $R=8$ up to $16$-hop). 
To isolate the effect of the training iteration strategy, we train all models on the same data (up to $12$-hop) and evaluate extrapolation on $12$–$24$ hop tasks.

\begin{figure}[h]
    \centering
    \includegraphics[width=\linewidth, trim=0 2 0 0, clip]{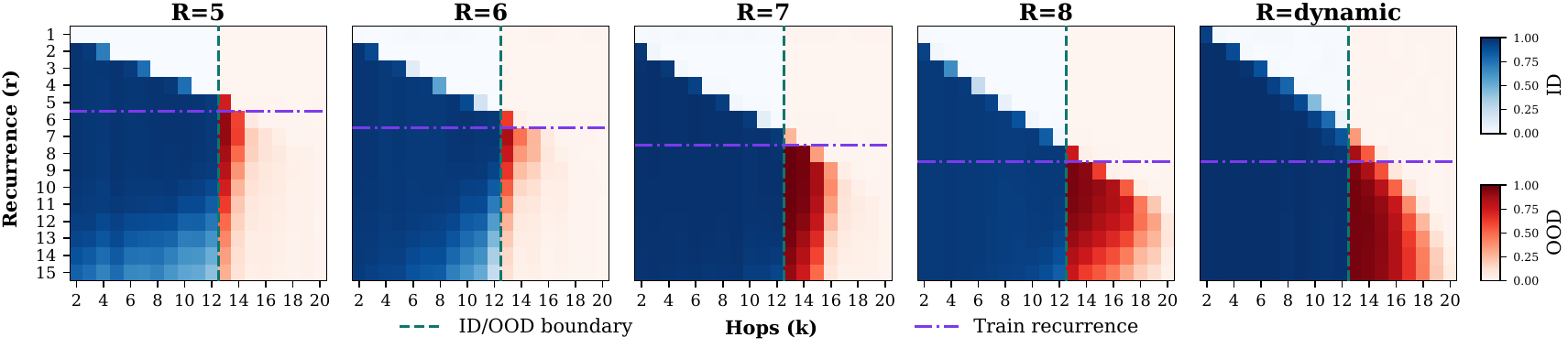}
    \caption{Accuracy of models trained with fixed recurrence $R \in \{5,6,7,8\}$ and dynamic recurrence on up to 12-hop samples.}
    \label{fig:extrapolation2}
\end{figure}

From Figure~\ref{fig:extrapolation2}, we first find that 
among fixed-iteration models, increasing the number of training-time iterations substantially improves extrapolation when scaling inference-time iterations (e.g.\ $R=6$ extrapolates up to $14$-hop, while $R=8$ reaches $19$-hop). 
Second, under the same training data, the dynamic iteration strategy achieves comparable extrapolation performance to the $R=8$ model (both reaching $19$-hop). 
This contrasts with Figure~\ref{fig:extrapolation}, where the dynamic strategy generalizes to significantly higher complexity than $R=8$.

These results suggest that the maximum number of iterations used during training determines the extrapolation range (i.e.\ how far beyond the training complexity the model can generalize), while dynamic iteration effectively exploits this range, since it enables a larger learnable recursion depth.
Hence, when training data is sufficiently complex, dynamic iteration should be preferred over fixed strategies with the same maximum iteration budget.
We provide further evidence of this through matched-data experiments in Appendix~\ref{app:matched_data}.

\paragraph{Scaling inference-time iterations is limited by overthinking.}
Despite strong generalization performance, we observe performance degradation when using excessively large inference-time iterations. 
For example, increasing iterations beyond 15 for the R=dynamic does not improve OOD performance (Figure~\ref{fig:extrapolation}), a phenomenon known as \textit{overthinking} \citep{bansal2022end}. 
We study this along two key axes: (1) inference iterations, i.e.\ how increasing the inference iterations affects the model’s prediction confidence, and (2) task complexity, i.e.\ how increasing task complexity impacts this confidence.

\begin{figure}[h]
    \centering
    \includegraphics[width=\linewidth, trim=0 10 0 0, clip]{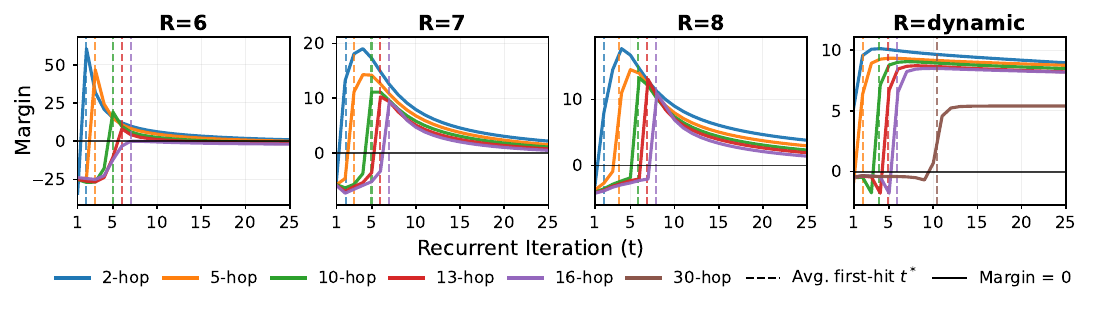}
    \caption{Average logit margin across recurrent iterations for fixed-recurrence ($R=6,7,8$) and dynamic-recurrence models. Curves show different hop complexities, and dashed vertical lines indicate the average first-hit iteration $t^*$ (when the correct entity is predicted).}
    \label{fig:margin}
\end{figure}

In Figure~\ref{fig:margin}, we analyze the logit margin, defined as the difference between the logit of the correct entity and that of the strongest competing token, and make two observations.
First, across all models and tasks, the margin increases with inference iterations until reaching a peak, and then consistently declines as iterations continue. 
This suggests that overthinking arises as the iteration number grows, regardless of the task complexity.
Notably, the dynamic model exhibits a much slower margin decay compared to fixed-iteration models, indicating that dynamic iteration is more robust to overthinking.
Second, the peak margin decreases as task complexity increases across all models. 
This implies that for more complex tasks, the model’s predictions are inherently less confident and therefore more susceptible to degradation under additional iterations. 
As a result, the benefit of increasing inference iterations diminishes for highly complex tasks.
Note that although previous work mitigates overthinking by injecting inputs information in every iteration \citep{bansal2022end, geiping2025scaling}, we find that such methods do not resolve the issue in implicit reasoning.

\begin{wrapfigure}{r}{0.35\textwidth}
    \vspace{-0.2em}
    \centering
    \includegraphics[width=0.33\textwidth]{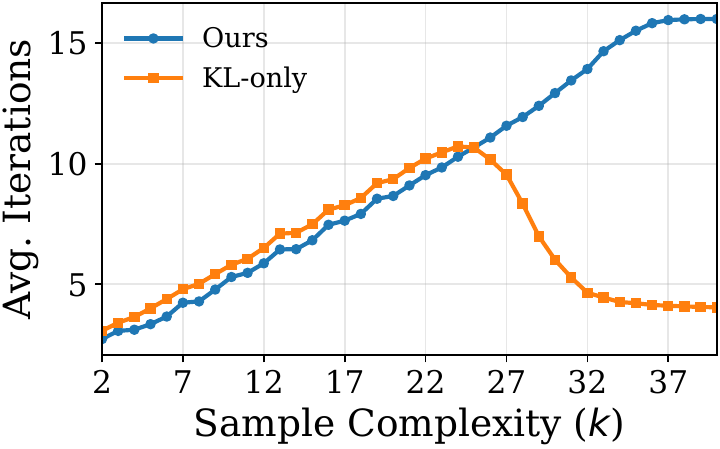}
    \vspace{-0.8em}
    \caption{Average recurrent iterations using adaptive halting vs. sample complexity.}
    \label{fig:adaptive-iterations-vs-hops}
    \vspace{-1.0em}
\end{wrapfigure}

\paragraph{Adaptive halting improves inference efficiency.} 
We flexibly halt recurrence in order to allocate compute proportionate to input complexity.
Prior work \citep{geiping2025scaling} proposes adaptive halting based on the output distribution $p_t(\cdot \mid x)$, terminating when the change between successive iterations becomes small, measured by $\mathrm{KL}\bigl(p_t ,|, p_{t-1}\bigr) < \varepsilon_{\mathrm{KL}}$.
In our setting, we find that this criterion often halts the recurrence prematurely, leading to suboptimal performance.
To address this, we additionally incorporate the entropy of the output distribution, $H(p_t)$, and only stop when both the divergence is small, and the prediction is confident, i.e.,

\[
\mathrm{KL}\!\bigl(p_t \,\|\, p_{t-1}\bigr) < \varepsilon_{\mathrm{KL}}
\qquad\text{and}\qquad
H(p_t) < H_{\mathrm{thresh}}.
\]

\begin{figure}[h]
    \centering
    \includegraphics[width=0.7\linewidth]{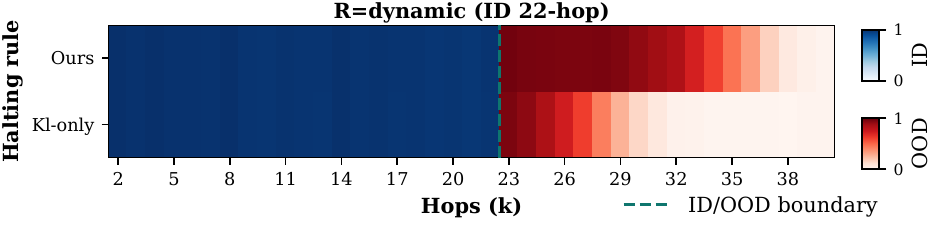}
    \caption{Comparison of adaptive halting based on KL-divergence and entropy (ours) versus KL-divergence alone \citep{geiping2025scaling}.}
    \label{fig:dynamic-adaptive-rows}
\end{figure}

In Figure~\ref{fig:adaptive-iterations-vs-hops}, we plot the number of iterations using both methods against the hop count ($k$), demonstrating that ours results in better allocation of inference-time compute in accordance with task complexity. While the output distribution may change very little across iterations, the entropy is still high, indicating that the model remains uncertain despite this apparent convergence. We compare results using the two methods in Figure~\ref{fig:dynamic-adaptive-rows}. In our setting, combining KL divergence with entropy therefore provides a more reliable halting signal and yields better overall results.

\section{Conclusion}
We study whether recurrent-depth transformers can compositionally use parametric knowledge for implicit multi-hop reasoning, a task that is challenging for vanilla transformers.
Through controlled experiments on models trained from scratch, we show that recurrent-depth transformers successfully address both compositional generalization challenges.
Systematic generalization emerges through a three-stage grokking dynamic, as the model transitions from rote memorization to generalizable solutions.
Depth extrapolation is enabled by scaling inference-time compute, where additional iterations allow for greater reasoning depth, although latent overthinking limits performance on highly complex tasks.
Overall, our results highlight recurrent-depth transformers as a promising architecture for compositional reasoning over parametric knowledge.

\section*{Acknowledgments}
The authors of this paper express their gratitude to Boshi Wang for valuable discussions on this work, and for motivating the permutation-based task formulation for depth exploration described in Appendix~\ref{app:permutation}. We also thank Yupei Du for valuable feedback on this paper, and the OSU NLP and DMRL groups for their thoughtful comments.

This research was supported in part by NSF CAREER \#1942980, Schmidt Sciences, Coefficient Giving (formerly Open Philanthropy) and Ohio Supercomputer Center \citep{Ohio_Supercomputer_Center1987-dl}. The views and conclusions contained herein are those of the authors and should not be interpreted as representing the official policies, either expressed or implied, of the U.S. government. The U.S. Government is authorized to reproduce and distribute reprints for Government purposes notwithstanding any copyright notice herein.

\bibliography{colm2026_conference}
\bibliographystyle{colm2026_conference}

\appendix
\section{Limitations}

Through our paper, we evaluate vanilla and recurrent-depth transformers on a single family of implicit reasoning problems. While we cleanly isolate systematic generalization and depth exploration in compositional reasoning, we do not cover the entire range of reasoning problems relevant to modern language models \citep{huang2023towards}. In order to maintain a controlled setting and properly study the behavior of recurrent-depth models, our experiments are intentionally small-scale and focused. This potentially limits direct extrapolation of results to modern LLMs whose performance and behavior is shaped by many additional factors like tokenizer choices, heterogeneous internet-scale data, pretraining data, post-training procedures and optimization at a much larger scale. Althought we study somewhat larger models in Appendix~\ref{app:ratios}, these are still significantly limited in scale relative to frontier LLMs today.

Additionally, our task formulation abstracts away many aspects of real-world language use. Inputs are presented in a highly structured form with dedicated entity and relation tokens with a very limited vocabulary. This removes challenges that might arise from surface-form variation, underspecification, distractor information, and distribution shift in natural language. Therefore, we do not claim complete and immediate transfer of positive results to LLMs trained with recurrent-depth at scale. However, we hope to isolate architectural effects in a controlled setting and use them to inform future designs for more robust implicit reasoning in future LLMs.

\section{Dataset}
\label{app:permutation}
To investigate how our implicit reasoning models scale to deeper $k$-hop composition tasks, we design a dataset from a permutation-based knowledge graph. In this setup, we found that models can sometimes achieve high accuracy via \emph{shortcuts} rather than performing the intended multi-step traversal. For large $k$, the final tail entity $t$ can become nearly determined by a short suffix of the relation sequence. Thus, a model may learn a shallow mapping from the trailing relations to the answer instead of retrieving the intermediate entities. To overcome this, We first construct a set of atomic facts over $|E|=200$ entities and $|R|=10$ relations. For each relation $r \in R$, we sample a random permutation $\pi_r$ over the entity indices and define the atomic facts as
\[
\forall e_i \in E:\quad (e_i, r, e_{\pi_r(i)}).
\]
Thus, each relation acts as a bijection over the entity set. Since the out-degree is set to $d=|R|=10$, every entity has exactly one outgoing edge for each relation. This ensures that each relation has full coverage over entities. To generate a $k$-hop example, we sample a starting atomic fact and then iteratively extend it by following outgoing edges from the current tail entity. At each step, one outgoing relation-edge pair is selected uniformly at random. Intermediate entities are used only during construction and are not included in the final training example. 


\section{Training details and additional parameters}\label{app:training}

For most experiments, we use an embedding dimension of 768, 12 attention heads and a recurrent block of 4 transformer layers. However, in Appendix~\ref{app:ratios} we experiment with recurrent blocks of greater depth. Similarly, in Section~\ref{sec:sys_results} we compare with a vanilla transformer of larger depth (8) in our systematicity analysis. AdamW optimizer \citep{loshchilovdecoupled} with a learning rate of $10^{-4}$, weight decay of 0.01 and a linear warmup schedule of 2000 steps is used in all of our experiments. We use a batch size of 512 and 128 for the systematicity and extrapolation experiments respectively. In our adaptive halting method, we use fixed thresholds $\varepsilon_{\mathrm{KL}} = 0.01$ and $H_{\mathrm{thresh}} = 3.00$, and halt recurrence when both $\mathrm{KL}\!\bigl(p_t \,\|\, p_{t-1}\bigr) < \varepsilon_{\mathrm{KL}}$ and $H(p_t) < H_{\mathrm{thresh}}$.

\section{Other related work}\label{app:other_related_work}

\paragraph {Implicit Reasoning LLMs.} Weight-sharing in transformers has been used as a strategy for parameter-efficiency starting with Universal Transformers \citep{dehghani2018universal} and ALBERT \citep{Lan2020ALBERT:}. An earlier closely related architecture is the ladder model by \citet{ju2022staircase}, which repeatedly applies a shared Transformer core to the full input sequence, yielding recurrence in depth. Recent work revisits this idea as a mechanism for implicit reasoning and scaling computation at inference-time. \citet{csordas2024moeut} introduce Mixture-of-Experts Universal Transformers (MoEUT), combining recurrent layer sharing with sparse experts to improve the parameter-compute trade-off. Recurrent-depth models trained at LLM scale such as Huginn \citep{geiping2025scaling} and Ouro \citep{zhu2025scaling} have shown promise and often exceed the performance of larger, non-recurrent models on popular benchmarks. Another approach to latent reasoning, Coconut \citep{hao2025training} replaces discrete reasoning tokens with continuous hidden states, enabling reasoning in latent space rather than through explicit text similar to recurrent-depth models. Complementarily, \citet{li2026polar} introduce Program-of-Layers (PoLar), a lightweight predictor that dynamically skips or repeats segments of frozen pretrained LLM layers for each input, improving reasoning accuracy often with fewer executed layers.

\paragraph{Studies on Implicit Reasoning in Recurrent Depth LLMs.} There is a growing body of work analyzing the improved performance of models through mechanisms like recurrent-depth architectures that enable reasoning in latent space. At LLM scale, \citet{lu2025latent} study the internals of the Huginn-3.5B using logit lens by projecting the intermediate representations through an unembedding matrix or through the models specialized coda layers (coda-lens). They report little evidence of actual latent reasoning and inconsistencies in interpretability across recurrent blocks. Conversely, \citet{du2025latent} find that trajectories of hidden states or "latent thoughts" that lead to correct outcomes are distinguishable from those that lead to incorrect outcomes through certain metrics. The representations of correct trajectories have a higher entropy, anisotropy, and intrinsic dimension but a lower effective rank indicating that correct thinking processes carry richer information with less noise, and generate more expressive latent representations. Both works take a pretrained, generalist LLM (Huginn-3.5B) and focus on probing its latent computations.  

\paragraph{Compositional Generalization.} Many recent studies have examined the performance of transformer-based language models in compositional generalization, motivated by the view that such step-by-step reasoning is central to human intelligence \citep{Simon1971HumanPS} and a proxy for analyzing how models can learn internal mechanisms for combining facts instead of emitting long rationales in the form of CoT. Using 3 representative tasks, \citet{10.5555/3666122.3669203} demonstrate that transformers reduce compositional tasks to linearized subgraph matching that fails with increasing complexity. \citet{10.5555/3737916.3740933} show that transformers learn implicit 2-hop compositional generalization only through "grokking" and that even then systematicity (OOD generalization) is never observed. \cite{kohli-etal-2025-groundcocoa} construct a logically grounded dataset to benchmark on compositional generalization and conditional reasoning, and show that even frontier LLMs struggle on tasks of sufficient compositional complexity. Experiments by \citet{yao-etal-2025-language-models} indicate that transformers are capable of multi-hop ID generalization but each hop requires exponentially larger amounts of data and that this issue is partially mitigated through curriculum learning. They also show that intermediate entities are retrieved \textit{layerwise} and that, in order to achieve $k$-hop generalization, the numbers of layers needs to grow linearly in $k$. \citet{balesni2024two} finetune pretrained LLMs on synthetic or semi-synthetic composition tasks and find that the models are only able to achieve composition when the two synthetic facts co-occur in the same finetuning document or test-time prompt or when one of the hops is a natural fact in the pretraining corpus. On algorithmic compositional tasks, \citet{csordas2022neural} introduce the Neural Data Router, augmenting a weight-shared Transformer with copy gates and geometric attention to learn adaptive control flow and improve generalization to longer and deeper compositions.

\section{Causal analysis of systematic generalization}
\label{app:causal}

\begin{figure}[h]
    \centering
    \includegraphics[width=\linewidth]{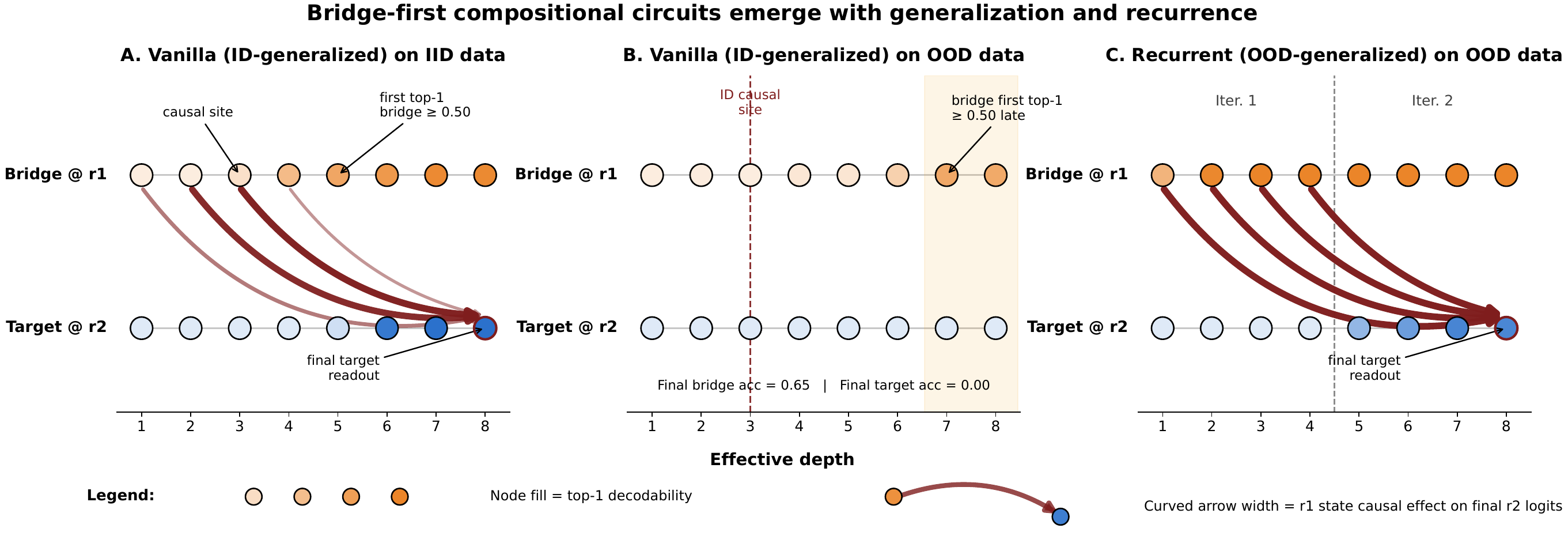}
    \caption{Bridge-mediated composition across effective depth in our systematicity setup.}
    \label{fig:causal}
\end{figure}

We additionally perform a causal analysis using activation patching together with logit-lens decodability, and present the results in Figure~\ref{fig:causal}. In the figure, the color shading of each node indicates decodability: for the top row, how strongly the bridge entity $b$ is decoded at token position $r_1$, and for the bottom row, how strongly the target entity $t$ is decoded at position $r_2$, across effective depth. The thickness of the curved arrows shows causal strength measured by activation patching: we corrupt the first-hop prefix $(h, r_1)$, restore the clean activation at position $r_1$ at a given effective depth, and measure how much the final target prediction is recovered. We use the term “causal site” in the figure to denote the layer at position $r_1$ that exhibits the strongest causal effect under activation patching.

Panel A shows that, on ID data, the bridge becomes decodable at $r_1$ in shallow layers (layers 5), and these layers exhibit causal effects on the final answer prediction. Panel B shows that, on OOD data, the vanilla model only recovers the bridge at much deeper layers (layer 7+), and the target never becomes decodable. This suggests that the model can access the OOD first-hop rule only too late to use it for the second-hop composition. Panel C shows that the recurrent-depth model alleviates this issue: all layers in the first iteration at position $r_1$ show causal effects on the final answer prediction, and subsequent recurrent iterations can reuse the same parametric rule to complete the second-hop composition.

\section{Experiments with default initialization}
\label{app:default-init}

\begin{figure}[h]
    \centering
    \includegraphics[width=\linewidth]{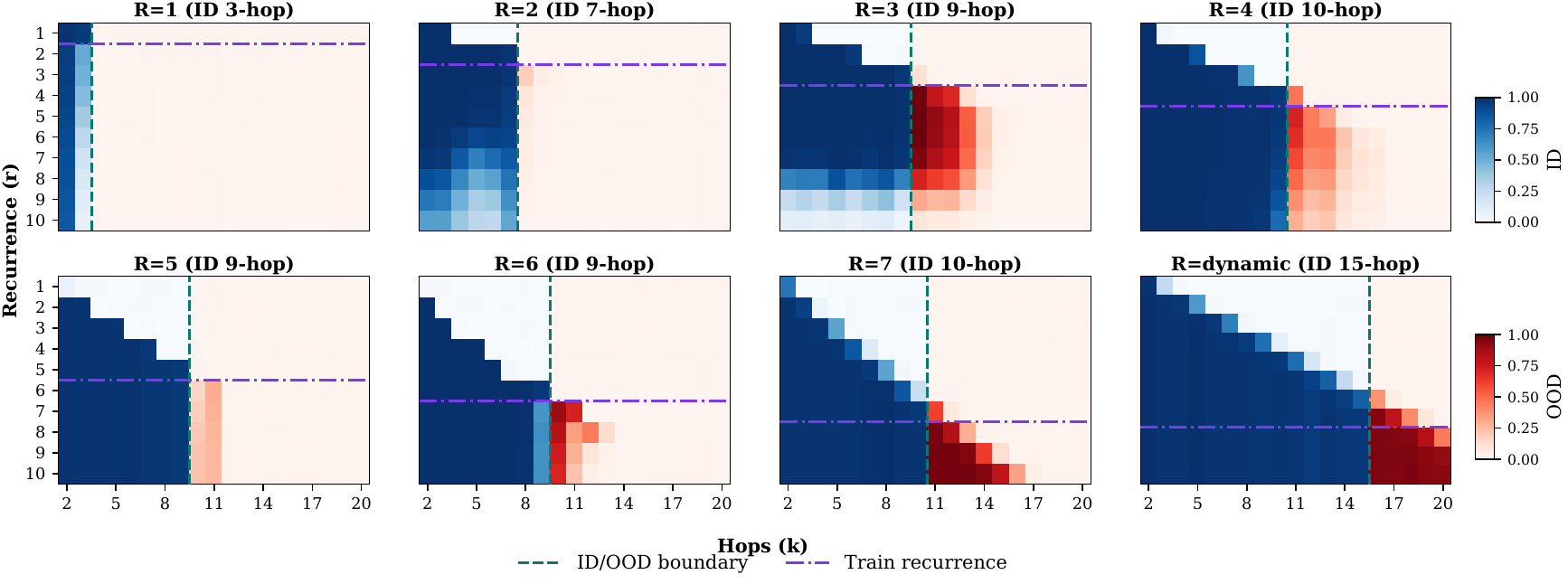}
    \caption{Results with default initialization for training recurrences $R \in \{1,\dots,7\}$ and dynamic recurrence.}
    \label{fig:default-init-main}
\end{figure}

Here we report results obtained using the default Gaussian initialization for the \texttt{c\_proj} matrices, rather than the zero-initialization described in Section~\ref{sec:model-design}. As observed from Figure~\ref{fig:default-init-main}, the results are mixed across training runs, and increasing recurrence does not yield a clear or consistent pattern of improved ID or OOD generalization as well as robustness to latent overthinking. Only the model trained with $R=7$, hows strong signs of inference-time scaling and robustness to latent overthinking. In the case of the model trained with $R=5$, there is no OOD generalization with increased recurrence at inference-time. For most of the other runs, we see performance degradation due to latent overthinking.

\begin{figure}[h]
    \centering
    \includegraphics[width=\linewidth]{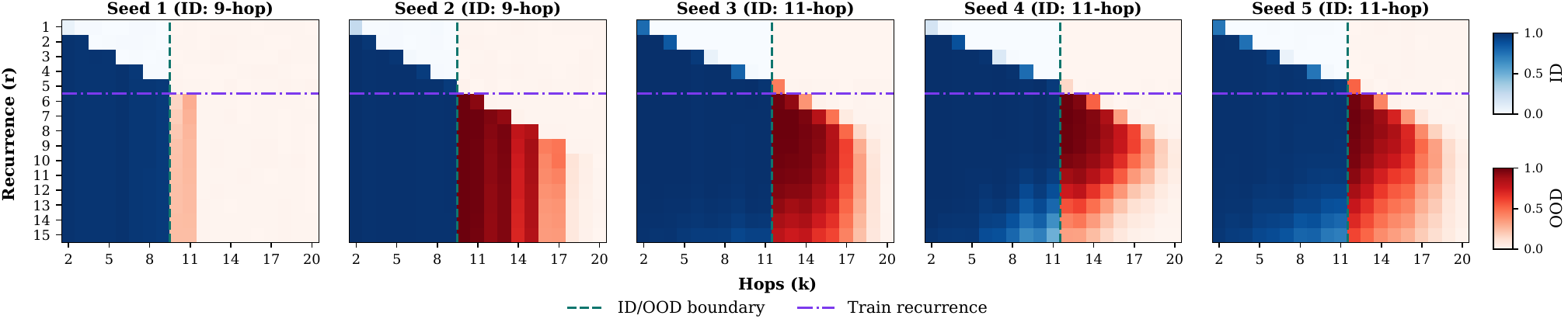}
    \caption{Five random-seed runs for the $R=5$ model with default initialization.}
    \label{fig:default-init-r5-seeds}
\end{figure}

To further examine this instability, we train the $r=5$ model with five different random seeds. As shown in Figure~\ref{fig:default-init-r5-seeds}, the resulting behaviors vary substantially across seeds. Seed 1 is not affected by latent overthinking, but it also shows no evidence of inference-time scaling by solving harder compositional examples with increased recurrence. Seed 2 is likewise stable, but does exhibit scaling with additional recurrence. Seeds 3, 4, and 5 show varying degrees of inference-time scaling, but also experience performance degradation as recurrence increases, consistent with latent overthinking.

\section{Experiments with larger vanilla transformers}
\label{app:isoflop_vanilla}

\begin{figure}[h]
    \centering \includegraphics[width=0.6\linewidth]{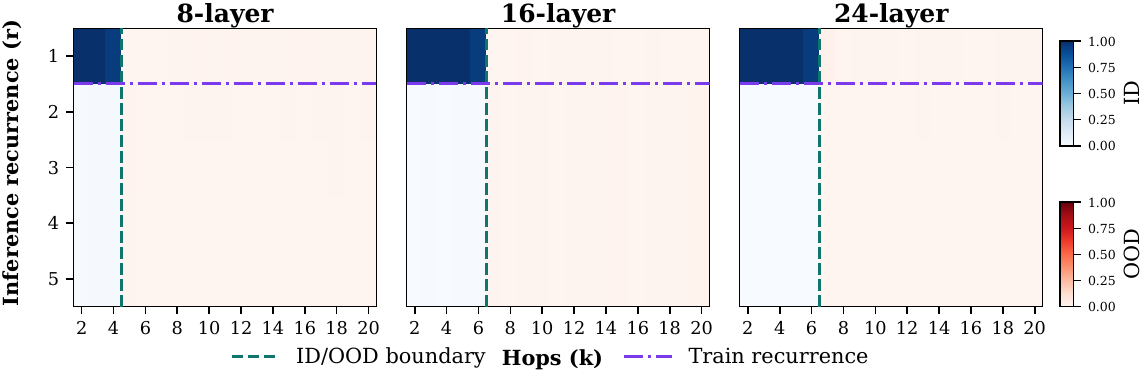}
    \caption{Depth extrapolation with larger vanilla transformers.}
    \label{fig:large_vanilla}
\end{figure}

We present additional results for larger vanilla transformers in Figure~\ref{fig:large_vanilla}. Specifically, we train 8-layer, 16-layer, and 24-layer vanilla transformers using the same setup as in Section~\ref{sec:exp_setup}. These results show that deeper vanilla models often achieve better ID generalization and can go beyond the 2-hop or 3-hop generalization observed in the 4-layer vanilla baseline. This is consistent with prior work showing that in-distribution generalization can improve with greater depth in vanilla transformers [1]. However, their ID generalization remains weaker than recurrent-depth models with the same effective depth. More importantly, the vanilla models still do not exhibit OOD generalization to more complex samples, and they lack the flexibility to scale computation according to sample complexity in the way recurrent-depth models can.

\section{Additional experiments with matched-hop data}
\label{app:matched_data}

\begin{figure}[h]
    \centering
    \includegraphics[width=\linewidth]{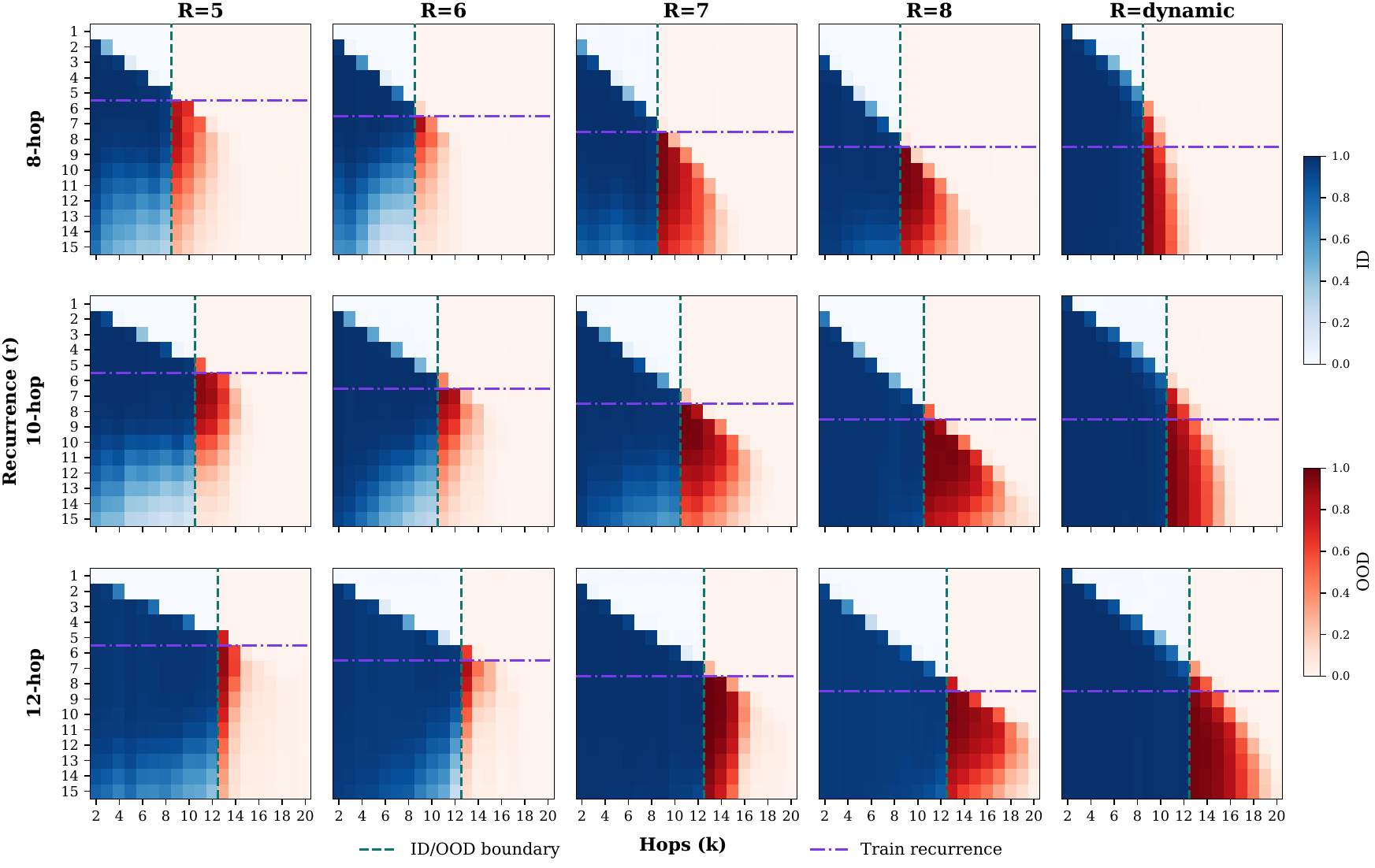}
    \caption{Matched-data experiments with varying train-time recurrence.}
    \label{fig:matched_large}
\end{figure}

We provide additional extrapolation results for models with different loop iterations trained on matched-hop data in Figure~\ref{fig:matched_large}. From top to bottom, the rows show results for models trained on data up to 8-hop, 10-hop, and 12-hop, while each column corresponds to a different training recurrence setting. The general trend is consistent with Figure~\ref{fig:extrapolation2}: models trained with higher recurrence achieve stronger OOD generalization through inference-time scaling.

\section{Results with different random seeds}\label{app:seeds}

\begin{figure}[h]
    \centering
    \includegraphics[width=0.9\linewidth]{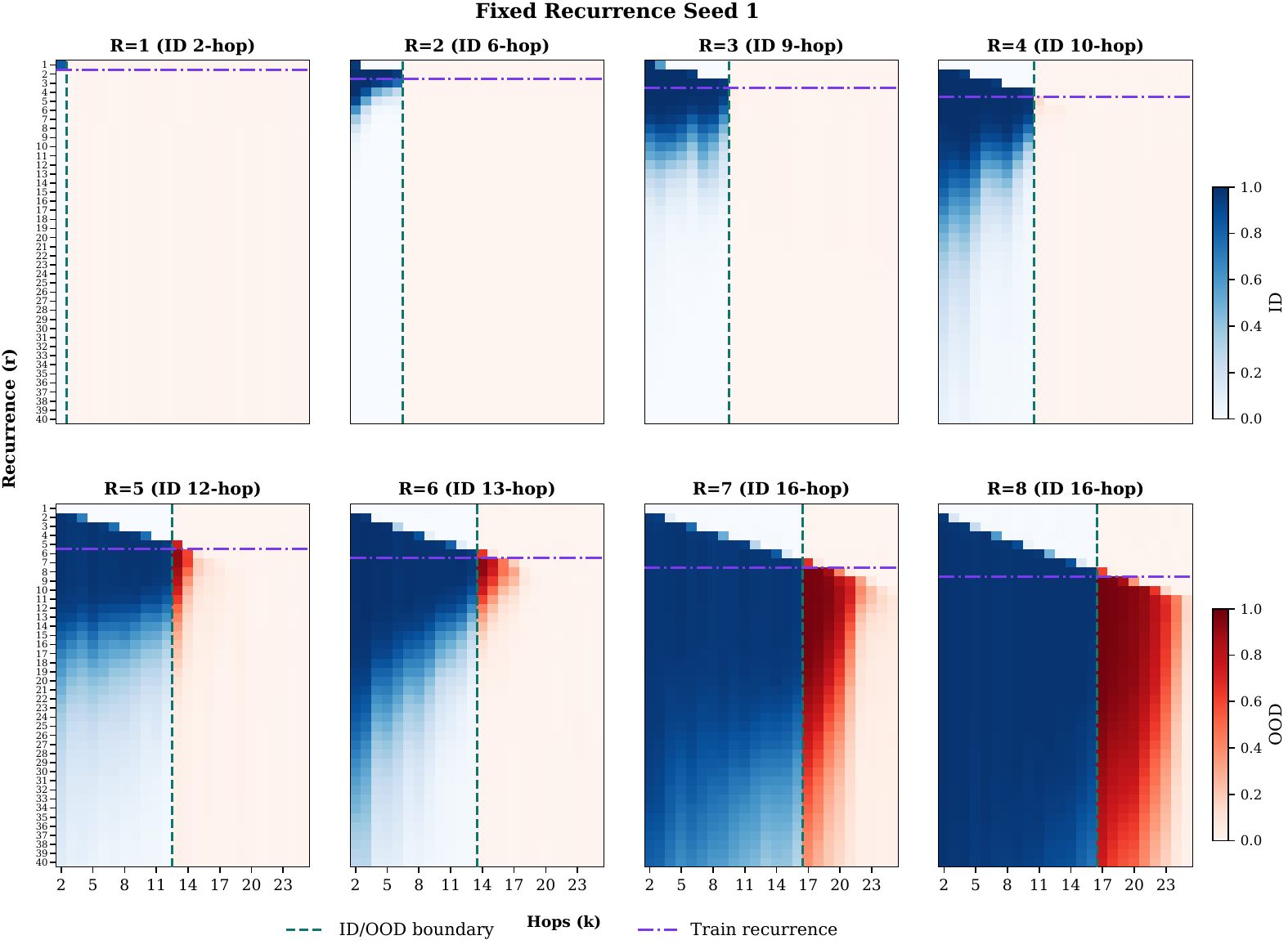}
    \caption{Results for Seed 1.}
    \label{fig:seed1}
\end{figure}

\begin{figure}[h]
    \centering
    \includegraphics[width=0.9\linewidth]{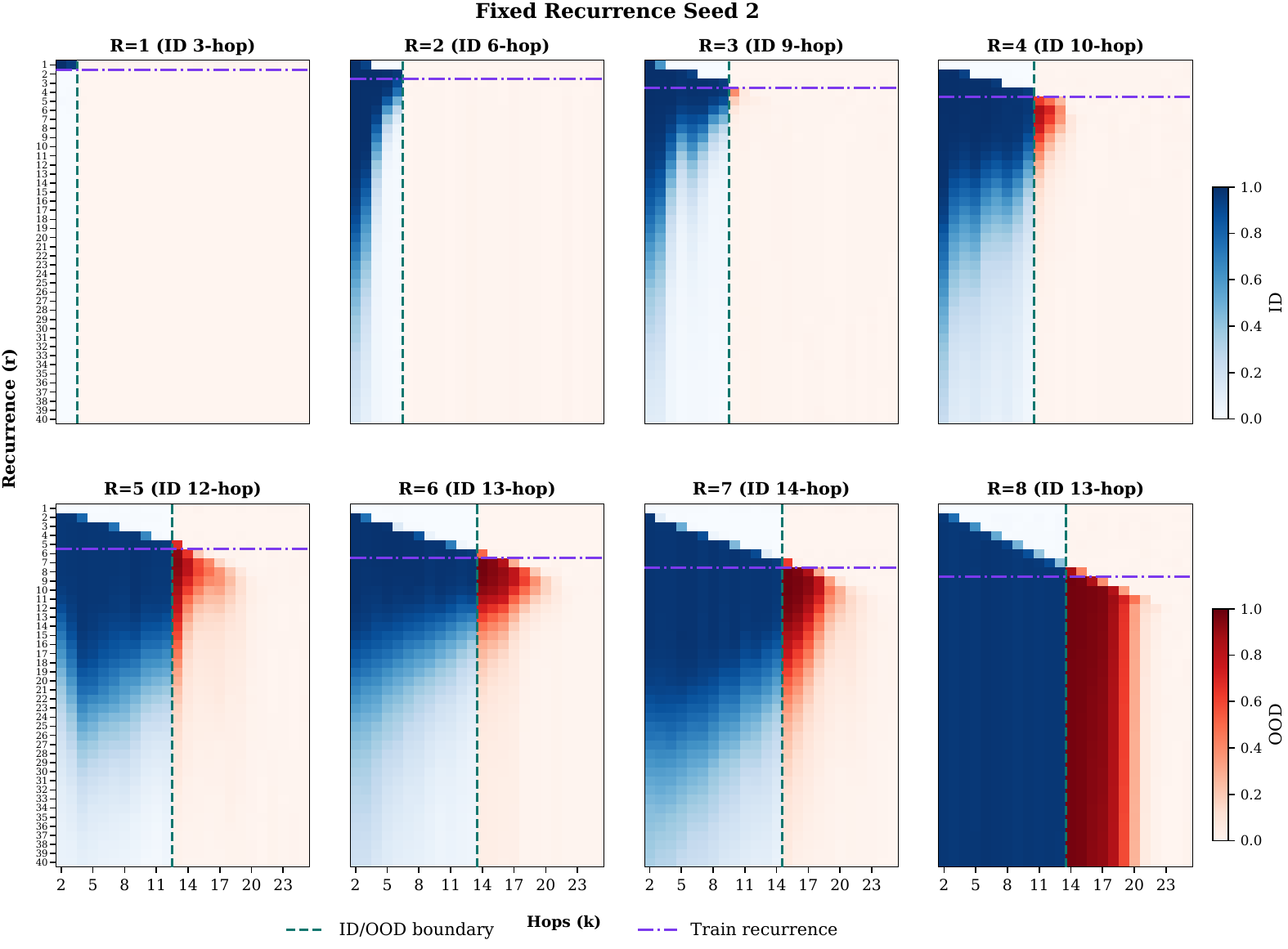}
    \caption{Results for Seed 2.}
    \label{fig:seed2}
\end{figure}

We run our fixed-recurrence models on 2 separate random seed initializations (including the run shown in Figure~\ref{fig:extrapolation}) and with to 40 recurrent iterations at inference time. On both runs, we notice the same general trend of increasing ID and OOD generalization with higher train-time recurrence as well as issues due to latent overthinking (with the sole exception being the model trained with $r=8$ in the second run). The results are presented in Figures~\ref{fig:seed1} \&~\ref{fig:seed2}.

We similarly plot the performance of the dynamic recurrence model across three different seeds and observe a generally high ID generalization and OOD extrapolation. Across all seeds, the dynamic recurrence model is robust to the latent overthinking phenomenon and adaptive halting ($r^*$) works reliably to stop recurrence when not necessary.

\begin{figure}[h]
    \centering
    \includegraphics[width=0.7\linewidth]{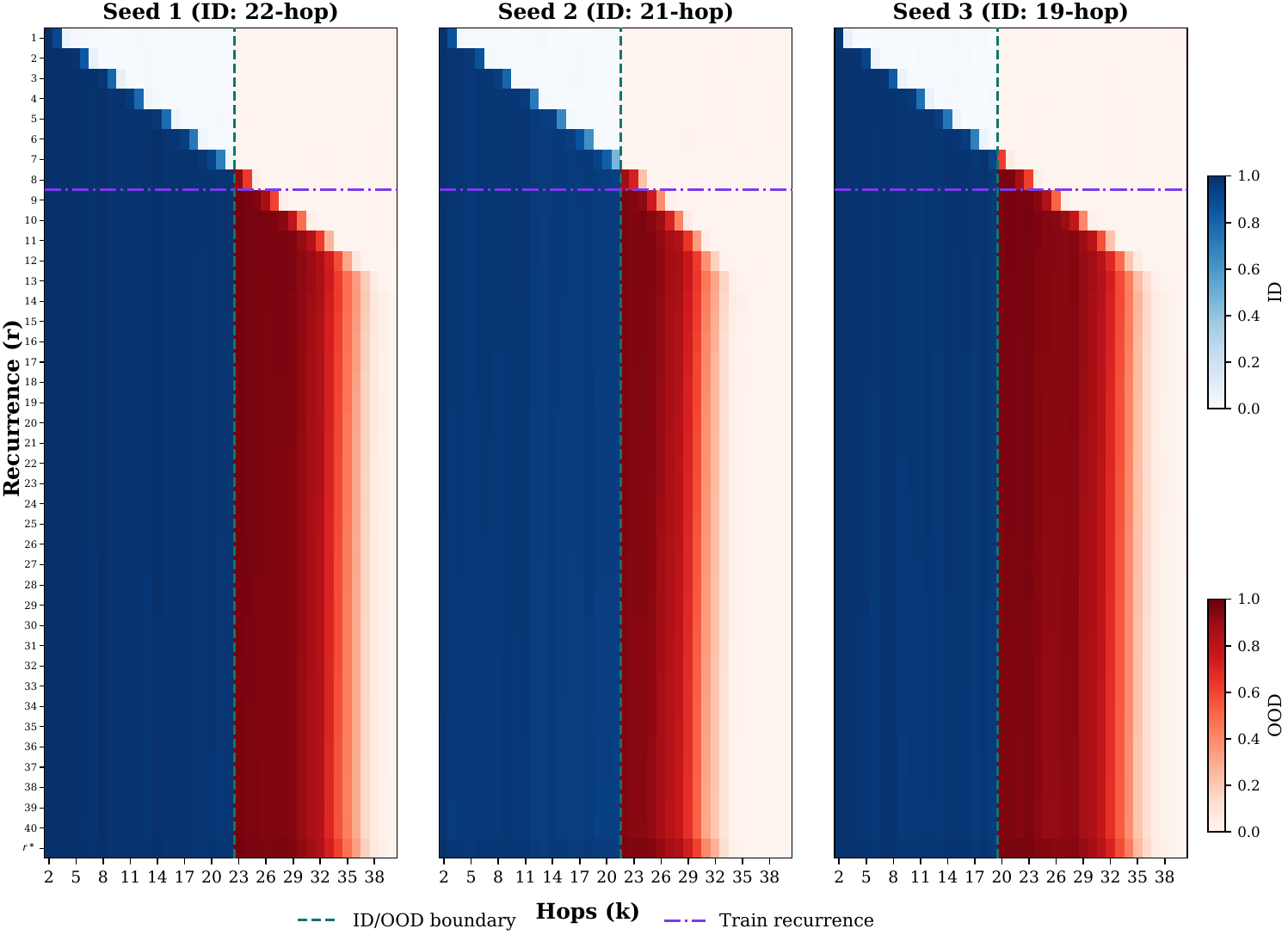}
    \caption{Dynamic recurrence with 3 difference random seed initializations.}
    \label{fig:dynamic_seeds}
\end{figure}

\section{Extrapolation with varying model size and dynamic recurrence}\label{app:ratios}

We carry out additional experiments to study how extrapolation behavior changes with model size and with training recurrence in our dynamic recurrence setting. In addition to our default 4-layer model, we train larger models with 6 and 8 layers. We also vary the maximum recurrence used during dynamic-recurrence training, considering settings with maximum train-time recurrence of 8, 12, and 16. For these, the recurrence at each training iteration is sampled from a Poisson distribution with means 4, 6, and 8 respectively, with a minimum recurrence of 2 in all cases.

\begin{figure}[h]
    \centering
    \includegraphics[width=\linewidth]{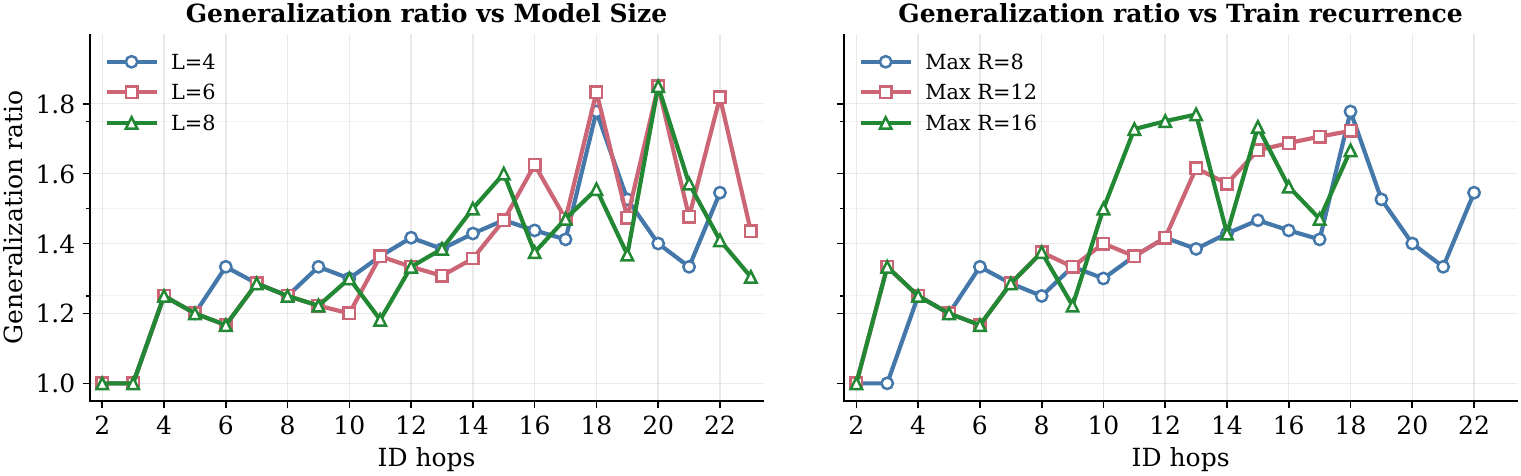}
    \caption{Generalization ratio with changing model size and train recurrence.}
    \label{fig:ratios}
\end{figure}

To summarize extrapolation under these settings, we plot the \emph{generalization ratio}, defined as the maximum hop complexity to which a model can generalize with at least 60\% accuracy with inference-time scaling, divided by the maximum hop complexity that the model has generalized to during training. This can be viewed as the ratio of maximum OOD generalization to maximum ID generalization. A ratio of 1 indicates no extrapolation beyond the level reached during training, while larger values indicate successful extrapolation to more complex compositions at test time.

Figure~\ref{fig:ratios} shows that models initially begin with a ratio close to 1. In particular, when models have only generalized to 2-hop composition during training, they do not yet extrapolate to more complex samples. As training progresses and the models have seen more complex compositions through our curriculum learning setup, the ratio increases. However, across both model-size variations and different maximum train-time recurrence settings, we do not observe a clear or consistent trend indicating that either larger models or larger train-time recurrence systematically improves this ratio.

\section{Illusion of very deep composition through shortcuts}
\label{app:shortcut-effects}

\begin{figure}[h]
    \centering
    \includegraphics[width=0.8\linewidth]{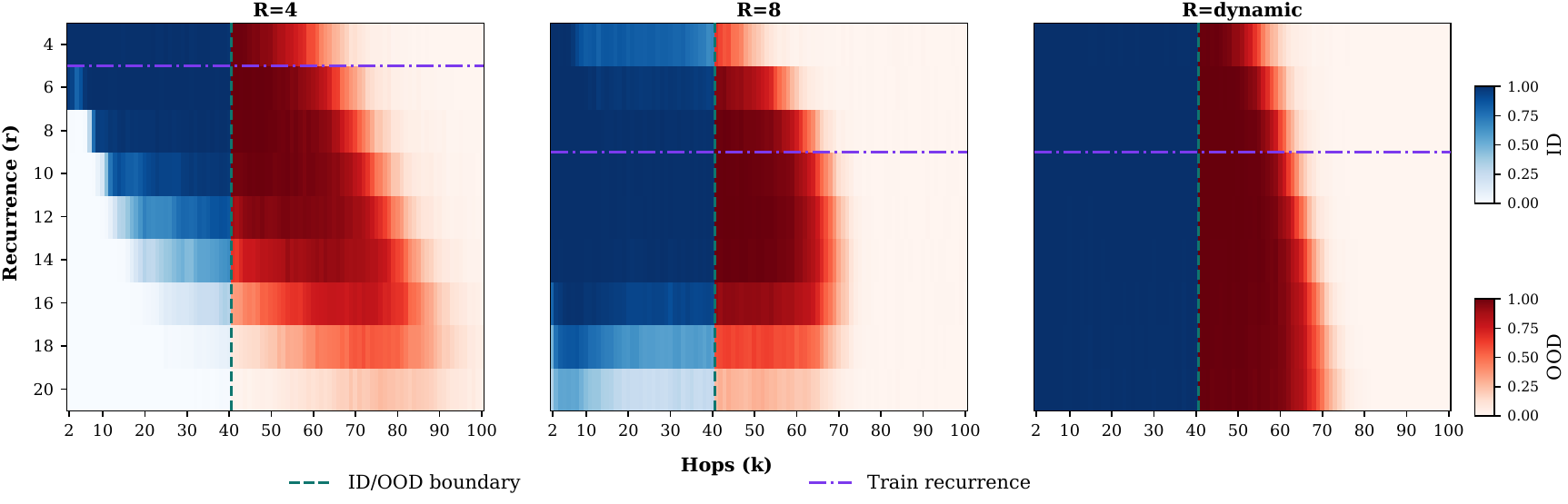}
    \caption{Apparent deep compositional generalization in the pre-permutation dataset.}
    \label{fig:shortcut2}
\end{figure}

\begin{figure}[h]
    \centering
    \includegraphics[width=0.9\linewidth]{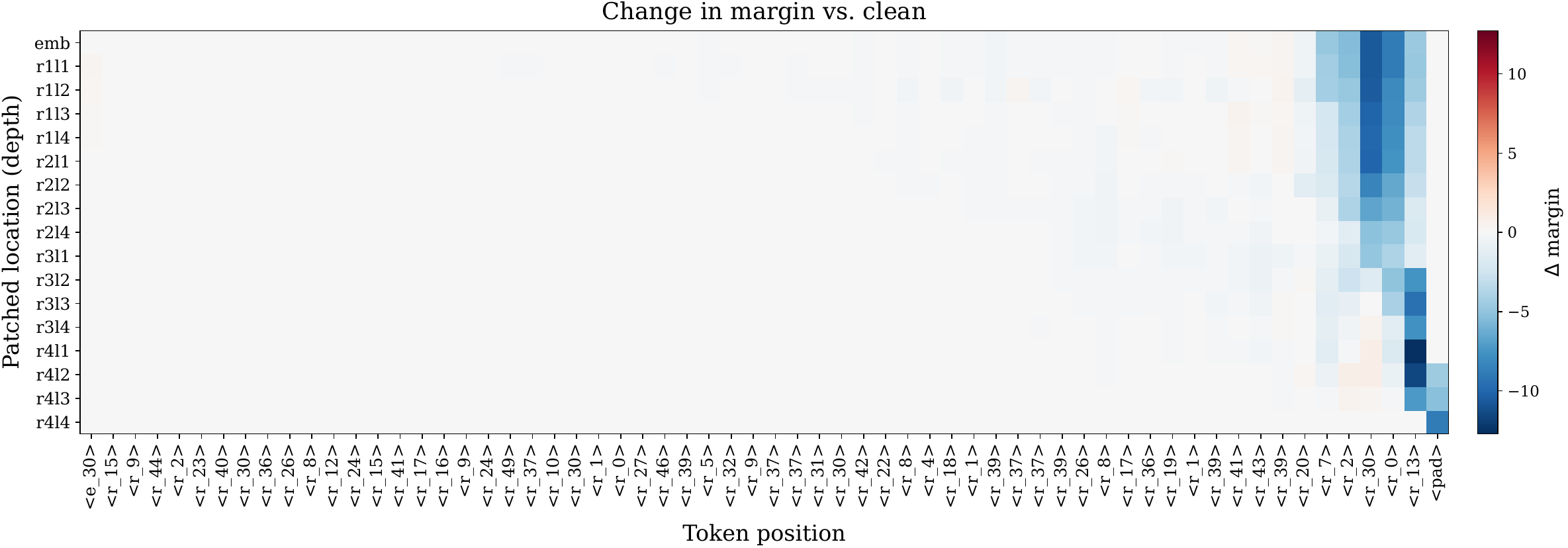}
    \caption{Activation-patching analysis on a 60-hop example. Each cell shows the change in logit margin for the correct answer, relative to the clean run, after replacing a single hidden state in the clean example with the corresponding hidden state from a random 60-hop example. Large changes are concentrated towards the end of the relation sequence, indicating shortcut-based prediction rather than genuine multi-hop retrieval.}
    \label{fig:shortcut1-deltas}
\end{figure}

Prior to adopting the permutation-based dataset construction described in Appendix~\ref{app:permutation}, we observed what appeared to be very deep compositional generalization. As shown in Figure~\ref{fig:shortcut2}, models trained on ID examples up to 40 hops achieved strong OOD performance even at 80 hops for models with $R=4$, $R=8$, and dynamic recurrence. This indicates that the model is able to resolve multiple hops or compositions in a single layer. To inspect this closely, we perform a causal activation-patching analysis. We first select a clean 60-hop test example whose relation sequence ends in a particular suffix, and a second random example of the same hop type. We then run both examples through the model and cache their hidden states. Next, for each token position and each recurrent iteration-layer location (as well as the embedding layer), we replace the hidden state in the clean run with the corresponding hidden state from the random run, and measure the resulting change in the logit margin of the correct final answer. Figure~\ref{fig:shortcut1-deltas} visualizes this change relative to the clean run.

This analysis reveals that the model is not solving the full compositional chain. Instead, the strongest causal effects are concentrated on a short suffix of the relation sequence. The final tail entity can often be inferred from trailing relations alone, without explicitly retrieving the intermediate entities. The model learns a shallow mapping from a suffix of the relation sequence to the answer, rather than performing full multi-hop composition. These observations motivate the permutation-based knowledge graph introduced in Appendix~\ref{app:permutation}. Under that construction, each relation acts as a permutation over entities, so the correct final entity cannot be recovered from a short suffix alone, and the model must compose the full sequence of relations to solve the task.

\end{document}